\documentclass[10pt,twocolumn,letterpaper]{article}

\usepackage{cvpr}
\usepackage{times}
\usepackage{epsfig}
\usepackage{graphicx}
\usepackage{amsmath}
\usepackage{amssymb}

\usepackage{enumitem}
\usepackage{amsfonts}
\usepackage{extarrows}
\usepackage{multirow}
\usepackage{rotating}
\usepackage{comment}
\usepackage{array}
\usepackage{booktabs}

\usepackage[pagebackref=true,breaklinks=true,letterpaper=true,colorlinks,bookmarks=false]{hyperref}

\cvprfinalcopy 

\def\cvprPaperID{2082} 

\ifcvprfinal\fi
\begin{document}

\title{Explaining AlphaGo: Interpreting Contextual Effects in Neural Networks}

\author{Zenan Ling$^{1,*}$, Haotian Ma$^{2}$, Yu Yang$^{3}$, Robert C. Qiu$^{1,4}$, Song-Chun Zhu$^{3}$, and Quanshi Zhang$^{1}$\\
$^{1}$Shanghai Jiao Tong University, \quad$^{2}$South China University of Technology,\\$^{3}$University of California, Los Angeles, \quad$^{4}$Tennessee Technological University}

\maketitle

\begin{abstract}
In this paper\footnote[1]{Quanshi Zhang is the corresponding author. Zenan Ling and Haotian Ma make equal contribution.}, we propose to disentangle and interpret contextual effects that are encoded in a pre-trained deep neural network. We use our method to explain the gaming strategy of the alphaGo Zero model. Unlike previous studies that visualized image appearances corresponding to the network output or a neural activation only from a global perspective, our research aims to clarify how a certain input unit (dimension) collaborates with other units (dimensions) to constitute inference patterns of the neural network and thus contribute to the network output. The analysis of local contextual effects \emph{w.r.t.} certain input units is of special values in real applications. Explaining the logic of the alphaGo Zero model is a typical application. In experiments, our method successfully disentangled the rationale of each move during the Go game.
\end{abstract}

\section{Introduction}

Interpreting the decision-making logic hidden inside neural networks is an emerging research direction in recent years. The visualization of neural networks and the extraction of pixel-level input-output correlations are two typical methodologies. However, previous studies usually interpret the knowledge inside a pre-trained neural network from a global perspective. For example, \cite{trust,shap,patternNet} mined input units (dimensions or pixels) that the network output is sensitive to; \cite{Interpretability} visualized receptive fields of filters in intermediate layers; \cite{CNNVisualization_1,CNNVisualization_2,CNNVisualization_3,FeaVisual,visualCNN_grad,visualCNN_grad_2} illustrated image appearances that maximized the score of the network output, a filter's response, or a certain activation unit in a feature map.

However, instead of visualizing the entire appearance that is responsible for a network output or an activation unit, we are more interested in the following questions.
\begin{itemize}
\item How does a local input unit contribute to the network output? Here, we can vectorize the input of the network into a high-dimensional vector, and we treat each dimension as a specific ``unit'' without ambiguity. As we know, a single input unit is usually not informative enough to make independent contributions to the network output. Thus, we need to clarify which other input units the target input unit collaborates with to constitute inference patterns of the neural network, so as to pass information to high layers.
\item Can we quantitatively measure the significance of above contextual collaborations between the target input unit and its neighboring units?
\end{itemize}

\textbf{Method:} Therefore, given a pre-trained convolutional neural network (CNN), we propose to disentangle contextual effects \emph{w.r.t.} certain input units.

\begin{figure*}
\centering
\includegraphics[width=0.95\linewidth]{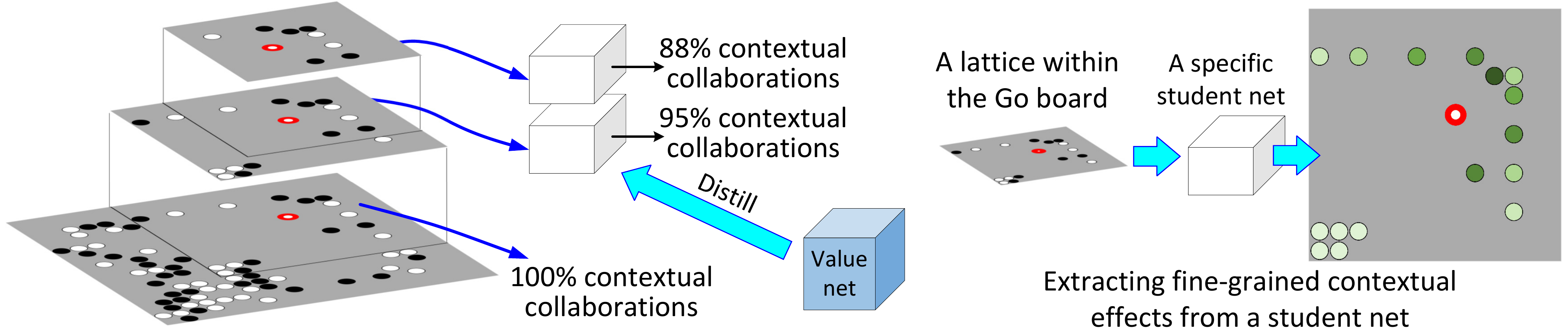}
\vspace{-5pt}
\caption{Explaining the alphaGo model. Given the state of the Go board and the next move, we use the alphaGo model to explain the rationale of the move. We first estimate a rough region of contextual collaborations \emph{w.r.t.} the current move by distilling knowledge from the value net to student nets that receive different regions of the Go board as inputs. Then, given a student net, we analyze fine-grained contextual collaborations within its region of the Go board. In this figure, we use a board state from a real Go game between humans for clarity.\vspace{-5pt}}
\label{fig:alphaGo}
\end{figure*}

\begin{figure*}
\centering
\includegraphics[width=0.99\linewidth]{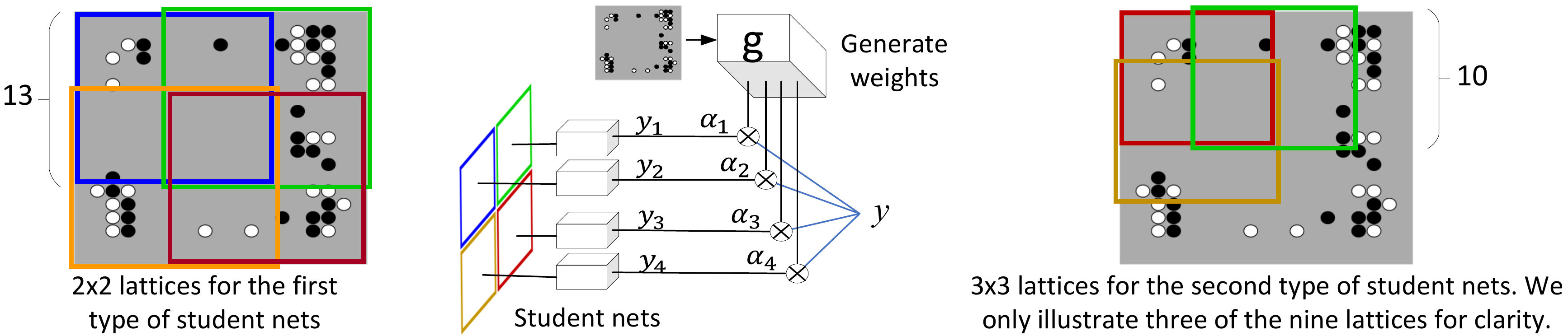}
\vspace{-5pt}
\caption{Division of lattices for two types of student nets. We distill knowledge from the value net into a mixture of four/nine student nets to approximate decision-making logic of the value net.\vspace{-5pt}}
\label{fig:student}
\end{figure*}

As shown in Fig.~\ref{fig:alphaGo}, we design two methods to interpret contextual collaborations at different scales, which are agnostic to the structure of CNNs. The first method estimates a rough region of contextual collaborations, \emph{i.e.} clarifying whether the target input unit mainly collaborates with a few neighboring units or most units of the input. This method distills knowledge from the pre-trained network into a mixture of local models (see Fig.~\ref{fig:student}), where each model encodes contextual collaborations within a specific input region to make predictions. We hope that the knowledge-distillation strategy can help people determine quantitative contributions from different regions. Then, given a model for local collaborations, the second method further analyzes the significance of detailed collaborations between each pair of input units, when we use the local model to make predictions on an image.

\textbf{Application, explaining the alphaGo Zero model:} The quantitative analysis of contextual collaborations \emph{w.r.t.} a local input unit is of special values in some tasks. For example, explaining the alphaGo model~\cite{alphaGo,alphaGoZero} is a typical application.

The alphaGo model contains a value network to evaluate the current state of the game---a high output score indicates a high probability of winning. As we know, the contribution of a single move (\emph{i.e.} placing a new stone on the Go board) to the output score during the game depends on contextual shapes on the Go board. Thus, disentangling explicit contextual collaborations that contribute to the output of the value network is important to understand the logic of each new move hidden in the alphaGo model.

More crucially, in this study, we explain the alphaGo Zero model~\cite{alphaGoZero}, which extends the scope of interests of this study from \textit{diagnosing feature representations of a neural network} to a more appealing issue \textit{letting self-improving AI teach people new knowledge}. The alphaGo Zero model is pre-trained via self-play without receiving any prior knowledge from human experience as supervision. In this way, all extracted contextual collaborations represent the automatically learned intelligence, rather than human knowledge.

As demonstrated in well-known Go competitions between the alphaGo and human players~\cite{alphaGo_Lee,alphaGo_Ke}, the automatically learned model sometimes made decisions that could not be explained by existing gaming principles. The visualization of contextual collaborations may provide new knowledge beyond people's current understanding of the Go game.

\textbf{Contributions} of this paper can be summarized as follows.\\
(i) In this paper, we focus on a new problem, \emph{i.e.} visualizing local contextual effects in the decision-making of a pre-trained neural network \emph{w.r.t.} a certain input unit.\\
(ii) We propose two new methods to extract contextual effects via diagnosing feature representations and knowledge distillation.\\
(iii) We have combined two proposed methods to explain the alphaGo Zero model, and experimental results have demonstrated the effectiveness of our methods.

\section{Related work}

Understanding feature representations inside neural networks is an emerging research direction in recent years. Related studies include 1) the visualization and diagnosis of network features, 2) disentangling or distilling network feature representations into interpretable models, and 3) learning neural networks with disentangled and interpretable features in intermediate layers.

\textbf{Network visualization:} Instead of analyzing network features from a global view~\cite{InformationBottleneck,InformationBottleneck2,CNNSpaceVisualization}, \cite{Interpretability} defined six types of semantics for middle-layer feature maps of a CNN, \emph{i.e.} \textit{objects}, \textit{parts}, \textit{scenes}, \textit{textures}, \textit{materials}, and \textit{colors}. Usually, each filter encodes a mixture of different semantics, thus difficult to explain.

Visualization of filters in intermediate layers is the most direct method to analyze the knowledge hidden inside a neural network. \cite{CNNVisualization_1,CNNVisualization_2,CNNVisualization_3,FeaVisual,CNNVisualization_6,CNNVisualization_7,attentionProp} showed the appearance that maximized the score of a given unit. \cite{FeaVisual} used up-convolutional nets to invert CNN feature maps to their corresponding images.

\textbf{Pattern retrieval:} Some studies retrieved certain units from intermediate layers of CNNs that were related to certain semantics, although the relationship between a certain semantics and each neural unit was usually convincing enough. People usually parallel the retrieved units similar to conventional mid-level features~\cite{MiddleLevel} of images. \cite{CNNSemanticDeep,CNNSemanticDeep2} selected units from feature maps to describe ``scenes''. \cite{ObjectDiscoveryCNN_2} discovered objects from feature maps.


\textbf{Model diagnosis and distillation:} Model-diagnosis methods, such as the LIME~\cite{trust}, the SHAP~\cite{shap}, influence functions~\cite{CNNInfluence}, gradient-based visualization methods~\cite{visualCNN_grad,visualCNN_grad_2}, and \cite{ExplainingArea} extracted image regions that were responsible for network outputs. \cite{additiveExplainer,explainer} distilled knowledge from a pre-trained neural network into explainable models to interpret the logic of the target network. Such distillation-based network explanation is related to the first method proposed in this paper. However, unlike previous studies distilling knowledge into explicit visual concepts, our using distillation to disentangle local contextual effects has not been explored in previous studies.

\textbf{Learning interpretable representations:} A new trend is to learn networks with meaningful feature representations in intermediate layers~\cite{LogicRuleNetwork,CNNCompositionality,Parsimonious} in a weakly-supervised or unsupervised manner. For example, capsule nets~\cite{capsule} and interpretable RCNN~\cite{InterRCNN} learned interpretable middle-layer features. InfoGAN~\cite{infoGAN} and $\beta$-VAE~\cite{betaVAE} learned meaningful input codes of generative networks. \cite{interpretableCNN} developed a loss to push each middle-layer filter towards the representation of a specific object part during the learning process without given part annotations.

All above related studies mainly focused on semantic meanings of a filter, an activation unit, a network output. In contrast, our work first analyzes quantitative contextual effects \emph{w.r.t.} a specific input unit during the inference process. Clarifying explicit mechanisms of how an input unit contributes to the network output has special values in applications.

\section{Algorithm}

In the following two subsections, we will introduce two methods that extract contextual collaborations \emph{w.r.t.} a certain input unit from a CNN at different scales. Then, we will introduce the application that uses the proposed methods to explain the alphaGo Zero model.

\subsection{Determining the region of contextual collaborations \emph{w.r.t.} an input unit}

Since the input feature usually has a huge number of dimensions (units), it is difficult to accurately discover a few input units that collaborate with a target input unit. Therefore, it is important to first approximate the rough region of contextual collaborations before the unit-level analysis of contextual collaborations, \emph{i.e.} clarifying in which regions contextual collaborations are contained.

Given a pre-trained neural network, an input sample, and a target unit of the sample, we propose a method that uses knowledge distillation to determine the region of contextual collaborations \emph{w.r.t.} the target input unit. Let $I\in{\bf I}$ denote the input feature (\emph{e.g.} an image or the state in a Go board). Note that input features of most CNNs can be represented as a tensor $I\in\mathbb{R}^{H\times W\times D}$, where $H$ and $W$ indicate the height of the width of the input, respectively; $D$ is the channel number. We clip different lattices (regions) $\Lambda_1,\Lambda_2,\ldots,\Lambda_{N}\in{\boldsymbol\Lambda}$ from the input tensor, and input units within the $i$-th lattice are given as $I_{\Lambda_{i}}\in\mathbb{R}^{h\times w\times D}$, $h\leq H,w\leq W$. Different lattices overlap with each other.

The core idea is that we use a mixture of models to approximate the function of the given pre-trained neural network (namely the \textit{teacher net}), where each model is a \textit{student net} and uses input information within a specific lattice $I_{\Lambda_{i}}$ to make predictions.
\begin{small}\begin{equation}
\hat{y}\approx\alpha_{1}\cdot y_{1}+\alpha_{2}\cdot y_{2}+\ldots+\alpha_{n}\cdot y_{n}
\end{equation}\end{small}
where $\hat{y}=f(I)$ and $y_{i}=f_{i}(I_{\Lambda_{i}})$ denote the output of the pre-trained teacher net $f$ and the output of the $i$-th student net $f_{i}$, respectively. $\alpha_{i}$ is a scalar weight, which depends on the input $I$. Because different lattices within the input are not equally informative \emph{w.r.t.} the target task, input units within different lattices make different contributions to final network output.

More crucially, given different inputs, the importance for the same lattice may also change. For example, as shown in \cite{visualCNN_grad_2}, the head appearance is the dominating feature in the classification of animal categories. Thus, if a lattice corresponds to the head, then this lattice will contribute more than other lattices, thereby having a large weight $\alpha_{i}$. Therefore, our method estimates a specific weight $\alpha_{i}$ for each input $I$, \emph{i.e.} $\alpha_{i}$ is formulated as a function of $I$ (which will be introduced later).

\textbf{Significance of contextual collaborations:} Based on the above equation, the significance of contextual collaborations within each lattice $\Lambda_{i}$ \emph{w.r.t.} an input unit can be measured as $s_{i}$.
\begin{equation}
\begin{split}
\Delta\hat{y}&\approx\!\!\!\!\!\!\!\!\!\!\!\!\!\!\!\underbrace{\alpha_{1}\cdot\Delta y_{1}}_{\textrm{Impacts from the first lattice }\Lambda_1}\!\!\!\!\!\!\!\!\!\!\!\!\!\!\!+\alpha_{2}\cdot\Delta y_{2}+\ldots+\alpha_{n}\cdot\Delta y_{n},\\
s_{i}&=\vert\alpha_{i}\cdot\Delta y_{i}\vert
\end{split}
\label{eqn:delta}
\end{equation}
where we revise the value of the target unit in the input and check the change of network outputs, $\Delta\hat{y}=f(I^{\textrm{new}})-f(I)$ and $\Delta y_{i}=f_{i}(I^{\textrm{new}}_{\Lambda_{i}})-f_{i}(I_{\Lambda_{i}})$. If contextual collaborations \emph{w.r.t.} the target unit mainly localize within the $i$-th lattice $\Lambda_{i}$, then $\alpha_{i}\cdot\Delta y_{i}$ can be expected to contribute the most to the change of $\hat{y}$.

We conduct two knowledge-distillation processes to learn student nets and a model of determining $\{\alpha_{i}\}$, respectively.

\textbf{Student nets:} The first process distills knowledge from the teacher net to each student net $f_{i}$ with parameters ${\boldsymbol\theta}_{i}$ based on the distillation loss $\min_{{\boldsymbol\theta}_{i}}\sum_{I\in{\bf I}}\Vert y_{I,i}-\hat{y}_{I}\Vert^2$, where the subscript $I$ indicates the output for the input $I$. Considering that $\Lambda_{i}$ only contains partial information of $I$, we do not expect $y_{I,i}$ to reconstruct $\hat{y}_{I}$ without any errors.

\textbf{Distilling knowledge to weights:} Then, the second distillation process estimates a set of weights ${\boldsymbol\alpha}=[\alpha_{I,1},\alpha_{I,2},\ldots,\alpha_{I,n}]$ for each specific input $I$. We use the following loss to learn another neural network $g$ with parameters ${\boldsymbol\theta}_{g}$ to infer the weight.
\begin{equation}
\begin{split}
&{\boldsymbol\alpha}=g(I),\qquad
\min_{{\boldsymbol\theta}_{g}}Loss({\boldsymbol\theta}_{g}),\\
&Loss({\boldsymbol\theta}_{g})=\sum_{I\in{\bf I}}\Vert\Delta\hat{y}_{I}-\sum_{i=1}^{N}\alpha_{I,i}\cdot\Delta y_{I,i}\Vert^2
\end{split}
\end{equation}

\subsection{Fine-grained contextual collaborations \emph{w.r.t.} an input unit}

In the above subsection, we introduce a method to distill knowledge of contextual collaborations into student nets of different regions. Given a student net, in this subsection, we develop an approach to disentangling from the student net explicit contextual collaborations \emph{w.r.t.} a specific input unit $u$, \emph{i.e.} identifying which input unit $v$ collaborates with $u$ to compute the network output.

We can consider a student net as a cascade of functions of $N$ layers, \emph{i.e.} $x^{(l)}=\phi_{l}(x^{(l-1)})$ (or $x^{(l)}=\phi_{l}(x^{(l-1)})+x^{(l-m)}$ for skip connections), where $x^{(l)}$ denotes the output feature of the $l$-th layer. In particular, $x^{(0)}$ and $x^{(n)}$ indicate the input and output of the network, respectively. We only focus on a single scalar output of the network (we may handle different output dimensions separately if the network has a high-dimensional output). If the sigmoid/softmax layer is the last layer, we use the score before the softmax/sigmoid operation as $x^{(n)}$ to simplify the analysis.

\subsubsection{Preliminaries, the estimation of quantitative contribution}
\label{sec:pre}

As preliminaries of our algorithm, we extend the technique of \cite{ContriProp} to estimate the quantitative contribution of each neural activation in a feature map to the final prediction. We use $C_{x}\in\mathbb{R}^{H_{l}\times W_{l}\times D_{l}}$ to denote the contribution distribution of neural activations on the $l$-th layer $x\in\mathbb{R}^{H_{l}\times W_{l}\times D_{l}}$. The score of the $i$-th element $C_{x_{i}}$ denotes the ratio of the unit $x_{i}$'s score contribution \emph{w.r.t.} the entire network output score. Because $x^{(n)}$ is the scalar network output, it has a unit contribution $C_{x^{(n)}}=1$. Then, we introduce how to back-propagate contributions to feature maps in low layers.

The method of contribution propagation is similar to network visualization based on gradient back-propagation~\cite{CNNVisualization_2,CNNVisualization_6}. However, contribution propagation reflects more objective distribution of numerical contributions over $\{x_{i}\}$, instead of biasedly boosting compacts of the most important activations.

Without loss of generality, in this paragraph, we use $o=\phi(x)$ to simplify the notation of the function of a certain layer. If the layer is a conv-layer or a fully-connected layer, then we can represent the convolution operation for computing each elementary activation score $o_{i}$ of $o$ in a vectorized form\footnote[2]{Please see the Appendix for details.} $o_{i}=\sum_{j}x_{j}w_{j}+b$. We consider $x_{j}w_{j}$ as the numerical contribution of $x_{j}$ to $o_{i}$. Thus, we can decompose the entire contribution of $o_{i}$, $C_{o_{i}}$, into elementary contributions of $x_{j}$, \emph{i.e.} $C_{o_{i}\rightarrow x_{j}}=C_{o_{i}}\cdot\frac{x_{j}w_{j}}{o_{i}+\max\{-b,0\}}$, which satisfies $C_{o_{i}\rightarrow x_{j}}\propto x_{j}w_{j}$ (see the appendix for details). Then, the entire contribution of $x_{j}$ is computed as the sum of elementary contributions from all $o_{i}$ in the above layer, \emph{i.e.} $C_{x_{j}}=\sum_{i}C_{o_{i}\rightarrow x_{j}}$.

A cascade of a conv-layer and a batch-normalization layer can be rewritten in the form of a single conv-layer, where normalization parameters are absorbed into the conv-layer\footnotemark[2]. For skip connections, a neural unit may receive contributions from different layers, $C_{x_{j}^{(l)}}=\sum_{i}C_{o_{i}^{(l+1)}\rightarrow x_{j}^{(l)}}+C_{x_{j}^{(l+m)}}$. If the layer is a ReLU layer or a Pooling layer, the contribution propagation has the same formulation as gradient back-propagations of those layers\footnotemark[2].

\subsubsection{The extraction of contextual collaborations}
\label{sec:contextual}

As discussed in \cite{Interpretability}, each neural activation $o_{i}$ of a middle-layer feature $o$ can be considered as the detection of a mid-level inference pattern. All input units must collaborate with neighboring units to activate some middle-layer feature units, in order to pass their information to the network output.

Therefore, in this research, we develop a method to\\
1. determine which mid-level patterns (or which neural activations $o_{i}$) the target unit $u$ constitutes;\\
2. clarify which input units $v$ help the target $u$ to constitute the mid-level patterns;\\
3. measure the strength of the collaboration between $u$ and $v$.

Let $o^{\textrm{bfr}}$ and $o$ denote the feature map of a certain conv-layer $o=f(x)$ when the network receives input features with the target unit $u$ being activated and the feature map generated without $u$ being activated, respectively. In this way, we can use $\vert o-o^{\textrm{bfr}}\vert$ to represent the absolute effect of $u$ on the feature map $o$. The overall contribution of the $i$-th neural unit $C_{o_{i}}$ depends on the activation score $o_{i}$, $C_{o_{i}}\propto\max\{o_{i},0\}$, where $\max\{o_{i},0\}$ measures the activation strength used for inference. The proportion of the contribution is affected by the target unit $u$ can be roughly formulated as $\mathcal{\tilde{C}}_{o}$.
\begin{equation}
\mathcal{\tilde{C}}_{o_{i}}=\left\{\begin{array}{ll}
C_{o_{i}}\frac{\vert o_{i}-o^{\textrm{bfr}}_{i}\vert}{o_{i}},&o_{i}>0\\
0,&\textrm{otherwise}
\end{array}\right.
\label{eqn:start}
\end{equation}
where $C_{o_{i}}=0$ and thus $\mathcal{\tilde{C}}_{o_{i}}=0$ if $o_{i}\leq0$, because negative activation scores of a conv-layer cannot pass information through the following ReLU layer ($o$ is not the feature map of the last conv-layer before the network output).

In this way, $\mathcal{\tilde{C}}_{o_{i}}$ highlights a few mid-level patterns (neural activations) related to the target unit $u$. $\mathcal{\tilde{C}}_{o}$ measures the contribution proportion that is affected by the target unit $u$. We can use $\mathcal{\tilde{C}}_{o}$ to replace $C_{o}$ and use techniques in Section~\ref{sec:pre} to propagate $\mathcal{\tilde{C}}_{o}$ back to input units $x^{(0)}$. Thus, $\mathcal{\tilde{C}}_{x^{(0)}}$ represents a map of fine-grained contextual collaborations \emph{w.r.t.} $u$. Each element in the map $\mathcal{\tilde{C}}_{x^{(0)}_{j}}$ is given as $x^{(0)}_{j}$'s collaboration with $u$.

We can understand the proposed method as follows. The relative activation change $\frac{\vert o_{i}-o^{\textrm{bfr}}_{i}\vert}{o_{i}}$ can be used as a weight to evaluate the correlation between $u$ and the $i$-th activation unit (inference pattern). In this way, we can extract input units that make great influences on $u$'s inference patterns, rather than affect all inference patterns. Note that both $u$ and $v$ may either increase or decrease the value of $o_{i}$. It means that the contextual unit $v$ may either boost $u$'s effects on the inference pattern, or weaken $u$'s effects.

\subsection{Application: explaining the alphaGo Zero model}

We use the ELF OpenGo~\cite{ELFOpenGo2018,tian2017elf} as the implementation of the alphaGo Zero model. We combine the above two methods to jointly explain each move's logic hidden in the value net of the alphaGo Zero model during the game. As we know, the alphaGo Zero model contains a value net, policy nets, and the module of the Monte-Carlo Tree Search (MCTS). Generally speaking, the superior performance of the alphaGo model greatly relies on the enumeration power of the policy net and the MCTS, but the value net provides the most direct information about \textit{how the model evaluates the current state of the game}. Therefore, we explain the value net, rather than the policy net or the MCTS. In the ELF OpenGo implementation, the value net is a residual network with 20 residual blocks, each containing two conv-layers. We take the scalar output\footnote[3]{The value net uses the current state, as well as seven most recent states, to output eight values for the eight states. To simplify the algorithm, we take the value corresponding to the current state as the target value.} before the final (sigmoid) layer as the target value to evaluate the current state on the Go board.

Given the current move of the game, our goal is to estimate unit-level contextual collaborations \emph{w.r.t.} the current move. \textit{\emph{I.e.} we aim to analyze which neighboring stones and/or what global shapes help the current move make influences to the game.} We distill knowledge from the value net to student networks to approximate contextual collaborations within different regions. Then, we estimate unit-level contextual collaborations based on the student net.

\textbf{Determining local contextual collaborations:} We design two types of student networks, which receive lattices at the scales of $13\times 13$ and $10\times 10$, respectively. In this way, we can conduct two distillation processes to learn neural networks that encode contextual collaborations at different scales.

As shown in Fig.~\ref{fig:student}, we have four student nets $\{f_{i}|i=1,\ldots,4\}$ oriented to $13\times 13$ lattices. Except for the output, the four student nets have the same network structure as the value net. The four student nets share parameters in all layers. The input of a student net only has two channels corresponding to maps of white stones and black stones, respectively, on the Go board. We crop four overlapping lattices at the four corners of the Go board for both training and testing. Note that we rotate the board state within each lattice $I_{\Lambda_{i}}$ to make the top-left position corresponds to the corner of the board, before we input $I_{\Lambda_{i}}$ to the student net. The neural network $g$ has the same settings as the value net. $g$ receives a concatenation of $[I_{\Lambda_{1}},\ldots,I_{\Lambda_{4}}]$ as the input. $g$ outputs four scalar weights $\{\alpha_{i}\}$ for the four local student networks $\{y_{i}\}$. We learn $g$ via knowledge distillation.

Student nets for $10\times 10$ lattices have similar settings as those for $13\times 13$ lattices. We divide the entire Go board into $3\times 3$ overlapping $10\times 10$ lattices. Nine student nets encode local knowledge from nine local lattices. We learn another neural network $g$, which uses a concatenation of $[I_{\Lambda_{1}},\ldots,I_{\Lambda_{9}}]$ to weight for the nine local lattices.

Finally, we select the most relevant $10\times 10$ lattice and the most relevant $13\times 13$ lattice, via $\max_{i}s_{i}$, for explanation.

\textbf{Estimating unit-level contextual collaborations:} In order to obtain fine-grained collaborations, we apply the method in Section~\ref{sec:contextual} to explain two student nets corresponding to the two selected relevant lattices. We also use our method to explain the value net. We compute a map of contextual collaborations for each neural network and normalize values in the map. We sum up maps of the three networks together to obtain the final map of contextual collaborations $\hat{C}$.

\begin{figure*}[t]
\centering
\includegraphics[width=0.99\linewidth]{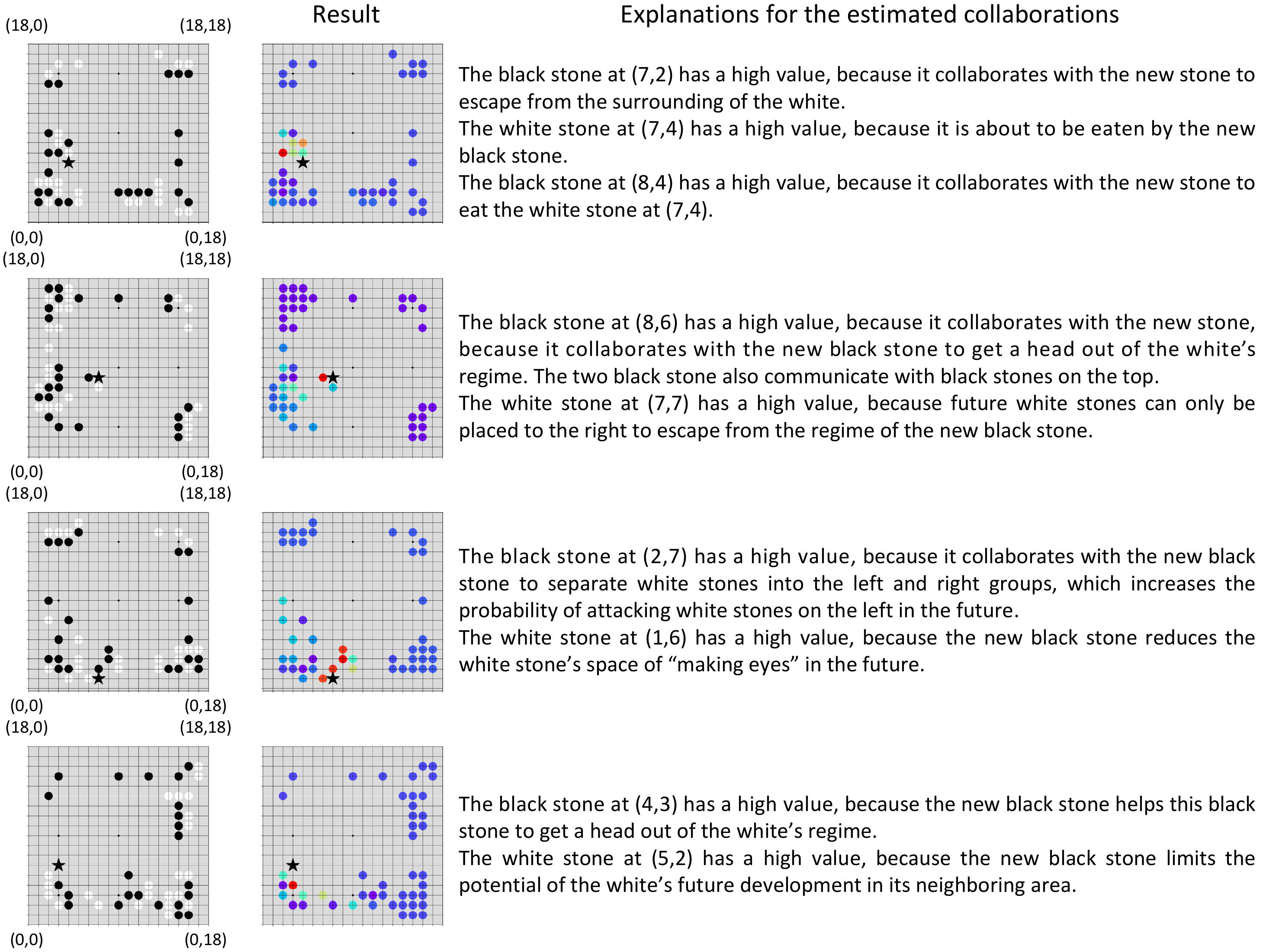}
\vspace{-5pt}
\caption{Significance of contextual collaborations \emph{w.r.t.} the new black stone (the black star). Go players provided possible explanations for contextual collaborations. The red/blue color indicates a significant/insignificant contextual collaboration. Please see the appendix for more results.}
\label{fig:analysis}
\end{figure*}

More specifically, given a neural network, we use the feature of each conv-layer to compute the initial $\mathcal{\tilde{C}}_{o}$ in Equation~(\ref{eqn:start}) and propagated $\mathcal{\tilde{C}}_{o}$ to obtain a map of collaborations $\mathcal{\tilde{C}}_{x^{(0)}}$. We sum up maps based on the 1st, 3rd, 5th, and 7th conv-layers to obtain the collaboration map of the network.


\section{Experiments}

In experiments, we distilled knowledge of the value network to student nets, and disentangled fine-grained contextual collaborations \emph{w.r.t.} each new move. We compared the extracted contextual collaborations and human explanations for the new move to evaluate the proposed method.

\subsection{Evaluation metric}

In this section, we propose two metrics to evaluate the accuracy of the extracted contextual collaborations \emph{w.r.t.} the new move. Note that \textit{considering the high complexity of the Go game, there is no exact ground-truth explanation for contextual collaborations. Different Go players usually have different analysis of the same board state.} More crucially, as shown in competitions between the alphaGo and human players~\cite{alphaGo_Lee,alphaGo_Ke}, the knowledge encoded in the alphaGo was sometimes beyond humans' current understanding of the Go game and could not be explained by existing gaming principles.

In this study, we compared the similarity between the extracted contextual collaborations and humans' analysis of the new move. The extracted contextual collaborations were just rough explanations from the perspective of the alphaGo. We expected these collaborations to be close to, but not exactly the same as human understanding. More specifically, we invited Go players who had obtained four-dan grading rank to label contextual collaborations. To simplify the metric, Go players were asked to label a relative strength value of the collaboration between each stone and the target move (stone), no matter whether the relationship between the two stones was collaborative or adversarial. Considering the double-blind policy, the paper will introduce the Go players if the paper is accepted.

Let $\Omega$ be a set of existing stones except for the target stone $u$ on the Go board. $p_{v}\geq0$ denotes the labeled collaboration strength between each stone $v\in\Omega$ and the target stone $u$. $q_{v}=\vert\hat{C}_{v}\vert$ is referred to as the collaboration strength estimated by our method, where $\hat{C}_{v}$ denotes the final estimated collaboration value on the stone $v$. We normalized the collaboration strength, $\hat{p}_{v}=p_{v}/\sum_{v'}p_{v'}$, $\hat{q}_{v}=q_{v}/\sum_{v'}q_{v'}$ and computed the Jaccard similarity between the distribution of $p$ and the distribution of $q$ as the similarity metric.

In addition, considering the great complexity of the Go game, different Go players may annotate different contextual collaborations. Therefore, we also required Go players to provide a subjective rating for the extracted contextual collaborations of each board state, \emph{i.e.} selecting one of the five ratings: 1-\textit{Unacceptable}, 2-\textit{Problematic}, 3-\textit{Acceptable}, 4-\textit{Good}, and 5-\textit{Perfect}.

\subsection{Experimental results and analysis}

Fig.~\ref{fig:analysis} shows the significance of the extracted contextual collaborations, as well as possible explanations for contextual collaborations, where the significance of the stone $v$'s contextual collaboration was reported as the absolute collaboration strength $q_{v}$ instead of the original score $\hat{C}_{v}$ in experiments. Without loss of generality, let us focus on the winning probability of the black. Considering the complexity of the Go game, there may be two cases of a positive (or negative) value of the collaboration score $\hat{C}_{v}$. The simplest case is that when a white stone had a negative value of $\hat{C}_{v}$, it means that the white stone decreased the winning probability of the black. However, sometimes a white stone had a positive $\hat{C}_{v}$. It may be because that this white stone did not sufficiently exhibit its power due to its contexts. Since the white and the white usually had a very similar number of stones in the Go board, putting a relatively ineffective white stone in a local region also wasted the opportunity of winning advantages in other regions in the zero-sum game. Similarly, the black stone may also have either a positive or a negative value of $\hat{C}_{v}$.

The Jaccard similarity between the extracted collaborations and the manually-annotated collaborations was 0.3633. Nevertheless, considering the great diversity of explaining the same game state, the average rating score that was made by Go players for the extracted collaborations was 3.7 (between 3-\textit{Acceptable} and 4-\textit{Good}). Please see the appendix for more results.

\section{Conclusion and discussions}

In this paper, we have proposed two typical methods for quantitative analysis of contextual collaborations \emph{w.r.t.} a certain input unit in the decision-making of a neural network. Extracting fine-grained contextual collaborations to clarify the reason why and how an input unit passes its information to the network output is of significant values in specific applications, but it has not been well explored before, to the best of our knowledge. In particular, we have applied our methods to the alphaGo Zero model, in order to explain the potential logic hidden inside the model that is automatically learned via self-play without human annotations. Experiments have demonstrated the effectiveness of the proposed methods.

Note that there is no exact ground-truth for contextual collaborations of the Go game, and how to evaluate the quality of the extracted contextual collaborations is still an open problem. As a pioneering study, we do not require the explanation to be exactly fit human logics, because human logic is usually not the only correct explanations. Instead, we just aim to visualize contextual collaborations without manually pushing visualization results towards human-interpretable concepts. This is different from some previous studies of network visualization~\cite{CNNVisualization_2,CNNVisualization_6} that added losses as the natural image prior, in order to obtain beautiful but biased visualization results. In the future, we will continue to cooperate with professional Go players to further refine the algorithm to visualize more accurate knowledge inside the alphaGo Zero model.

{\small
\bibliography{TheBib}
\bibliographystyle{ieee}
}

\newpage
\onecolumn
\appendix
\section*{Supplementary materials for the contribution propagation}

Let $o={\boldsymbol\omega}\otimes x+\beta$ denote the convolutional operation of a conv-layer. We can rewrite the this equation in a vectorized form as ${\bf o}=W{\bf x}+{\bf b}$, ${\bf o},{\bf b}\in\mathbb{R}^{1\times N}$, $W\in\mathbb{R}^{N\times N}$. For each output element $o_{i}$, $o_{i}=\sum_{j}x_{j}W_{ij}+b_{i}$. If the conv-layer is a fully-connected layer, then each element $W_{ij}$ corresponds to an element in ${\boldsymbol\omega}$. Otherwise, $W$ is a sparse matrix, \emph{i.e.} $W_{ij}=0$ if $o_{i}$ and $x_{j}$ are too far way to be covered by the convolutional filter.

Thus, we can write $o_{i}=\sum_{j}x_{j}w_{j}+b$ to simplify the notation. Intuitively, we can propagate the contribution of $o_{i}$ to its compositional elements $x_{j}$ based on their numerical scores. Note that we only consider the case of $o_{i}>0$, because if $o_{i}\leq0$, $o_{i}$ cannot pass information through the ReLU layer, and we obtain $C_{o_{i}}=0$ and thus $C_{o_{i}\rightarrow x_{j}}=0$. In particular, when $b\geq0$, all compositional scores just contribute an activation score $o_{i}-b$, thereby receiving a total contribution of $C_{o_{i}}\frac{o_{i}-b}{o_{i}}$. When $b<0$, we believe the contribution of $C_{o_{i}}$ all comes from elements of $\{x_{j}\}$, and each element's contribution is given a $C_{o_{i}}\cdot\frac{x_{j}w_{j}}{o_{i}-b}$. Thus, we get
\begin{equation}
C_{o_{i}\rightarrow x_{j}}=C_{o_{i}}\cdot\frac{x_{j}w_{j}}{o_{i}+\max\{-b,0\}}\nonumber
\end{equation}

When a batch-normalization layer follows a conv-layer, then the function of the two cascaded layers can be written as
\begin{equation}
\begin{split}
o_{i}=&\gamma\frac{(\sum_{j}x_{j}w_{j}+b)-\mu}{\sigma}+\beta'\\
=&\sum_{j}(\frac{\gamma w_{j}}{\sigma})x_{j}+[\frac{\gamma(b-\mu)}{\sigma}+\beta']
\end{split}
\nonumber
\end{equation}
Thus, we can absorb parameters for the batch normalization into the conv-layer, \emph{i.e.} $w_{j}\leftarrow\frac{\gamma w_{j}}{\sigma}$ and $b\leftarrow\frac{\gamma(b-\mu)}{\sigma}+\beta'$.

For ReLU layers and Pooling layers, the formulation of the contribution propagation is identical to the formulation for the gradient back-propagation, because the gradient back-propagation and the contribution propagation both pass information to neural activations that are used during the forward propagation.

\newpage
\section*{More results}
\textcolor{blue}{Considering the great complexity of the Go game, there do not exist ground-truth annotations for the significance of contextual collaborations. Different Go players may have different understanding of the same Go board state, thereby annotating different heat maps for the significance of contextual collaborations. More crucially, our results reflect the logic of the automatically-learned alphaGo Zero model, rather than the logic of humans.}

\textcolor{blue}{Therefore, in addition to manual annotations of collaboration significance, we also require Go players to provide a subjective evaluation for the extracted contextual collaborations.}

\includegraphics[width=0.7\linewidth]{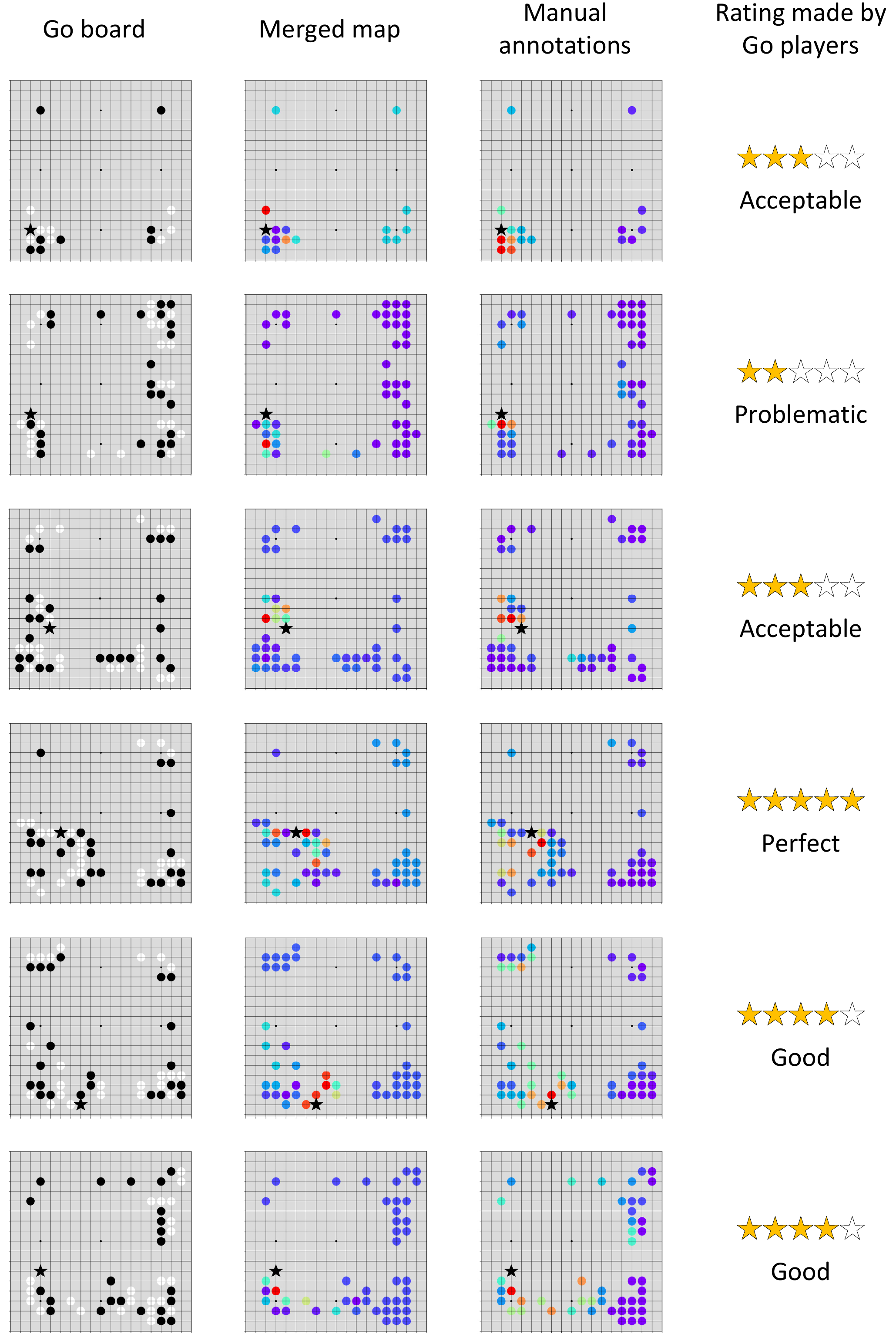}

We compared the extracted contextual collaborations at different scales (the second, third, fourth, and fifth columns) with annotations made by Go players.

\includegraphics[width=0.7\linewidth]{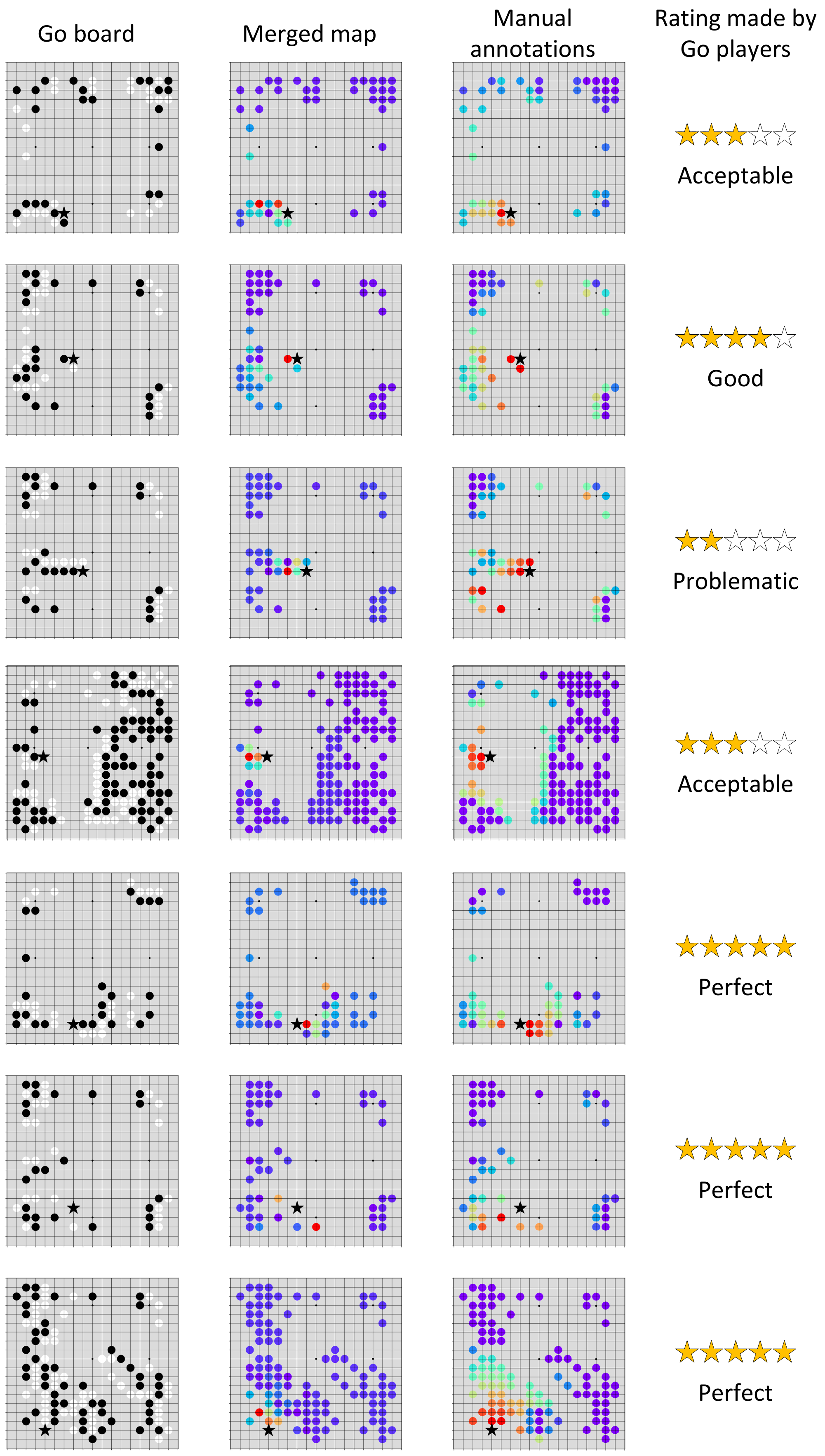}

We compared the extracted contextual collaborations at different scales (the second, third, fourth, and fifth columns) with annotations made by Go players.

\newpage
\section*{Contextual collaborations of local regions}

\includegraphics[width=0.9\linewidth]{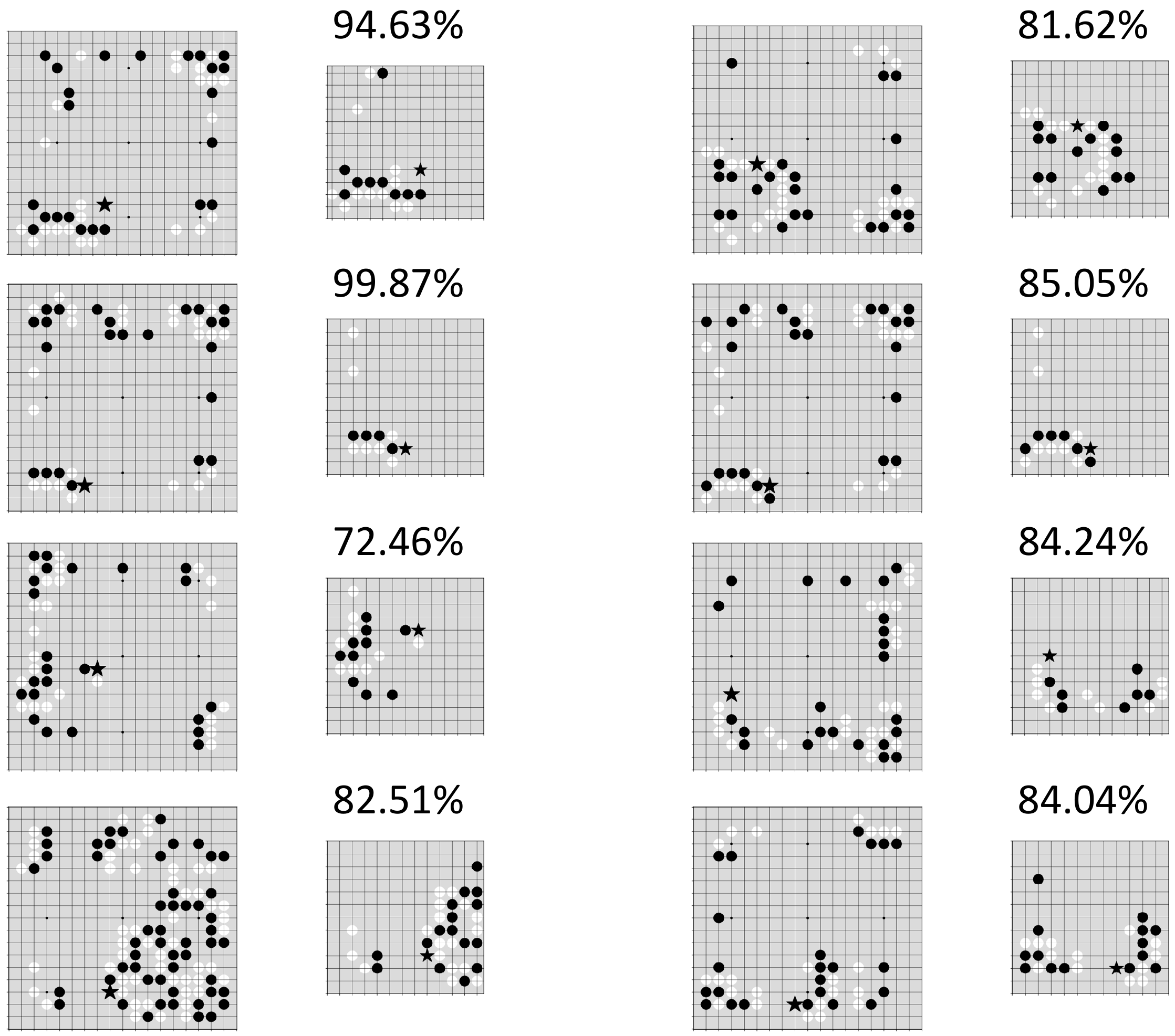}

We show the significance of contextual collaborations within a local lattice. The score for the $i$-th lattice is reported as $\frac{s_{i}}{\sum_{j}s_{j}}$.

\end{document}